%% file: localizing_moments_arxiv.tex
\documentclass[11pt,a4paper]{article}
\usepackage[hyperref]{emnlp2018}
\usepackage{times}
\usepackage{latexsym}
\usepackage{comment}
\usepackage{graphicx}
\usepackage{tabularx,ragged2e}
\usepackage{booktabs}
\usepackage{amsmath,amssymb,amsfonts} %
\usepackage{calligra}
\usepackage{bbm}
\usepackage{color}
\usepackage{url}

\newcommand{\Tau}{\mathrm{T}}

\definecolor{darkgreen}{RGB}{80,190,5}
\definecolor{darkblue}{RGB}{30,5,190}

\newcommand{\modelname}{MLLC}
\newcommand{\datasetname}{TEMPO}
\newcommand{\myparagraph}[1]{\vspace{3pt}\noindent{\bf #1}}

\aclfinalcopy %

\title{Localizing Moments in Video with Temporal Language}

\author{\textbf{Lisa Anne Hendricks}$^{1}$\thanks{\ \ Work done at Adobe during LAH's summer internship.},
\textbf{Oliver Wang}$^2$,
\textbf{Eli Shechtman}$^2$, \\
\textbf{Josef Sivic}$^{2,3*}$ , 
  \textbf{Trevor Darrell}$^1$,
\textbf{Bryan Russell}$^2$\\
$^1$ UC Berkeley, $^2$ Adobe Research, $^3$ INRIA \\
}
\date{}
\hypersetup{draft}
\begin{document}
\maketitle
\begin{abstract}
Localizing moments in a longer video via natural language queries is a new, challenging task at the intersection of language and video understanding.
Though moment localization with natural language is similar to other language and vision tasks like natural language object retrieval in images, moment localization offers an interesting opportunity to model temporal dependencies and reasoning in text.
We propose a new model that explicitly reasons about different temporal segments in a video, and shows that temporal context is important for localizing phrases which include temporal language.
To benchmark whether our model, and other recent video localization models, can effectively reason about temporal language, we collect the novel TEMPOral reasoning in video and language (\datasetname{}) dataset.
Our dataset consists of two parts: a dataset with real videos and template sentences (\datasetname{} - Template Language) which allows for controlled studies on temporal language, and a human language dataset which consists of temporal sentences annotated by humans (\datasetname{} - Human Language).

\end{abstract}

\input{introduction}
\input{related}

\input{method}
\input{dataset}
\input{results}
\input{acknowledgements}
\clearpage

\appendix
\section*{Appendix}
\input{supp}

\clearpage

\bibliography{emnlp2018}
\bibliographystyle{acl_natbib_nourl}

\end{document}

%% file: introduction.tex
\section{Introduction}

\begin{figure}[t]
\centering
\includegraphics[width=1.0\linewidth]{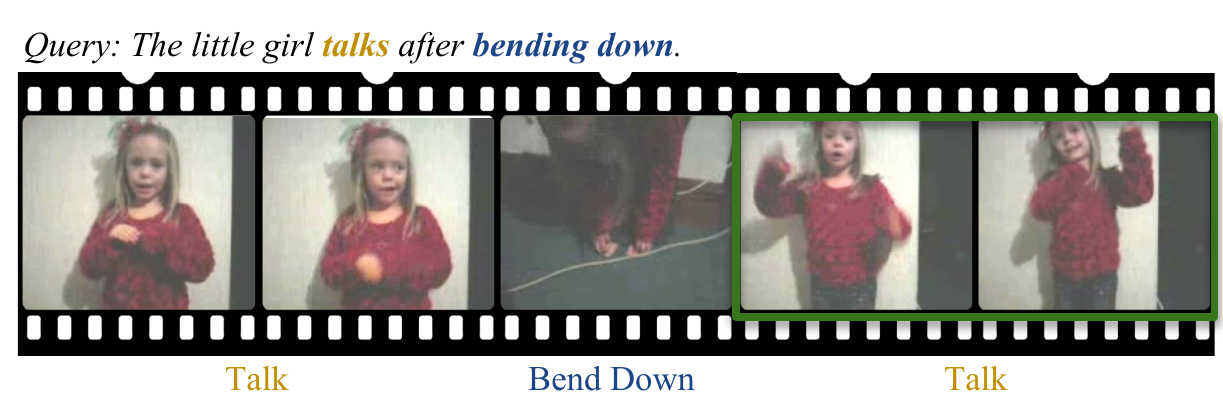}
\caption{We consider localizing video moments which include \textit{temporal language}.  To properly localize ``The little girl talks after bending down'' localization models must understand how the action ``talks'' relates to the action ``bend down.''  
}
\label{fig:teaser}
\vspace{-5mm}
\end{figure}

Consider the video and natural language query in Figure~\ref{fig:teaser} where we seek to localize the desired moment in the video specified by the query.
Queries like ``the girl bends down'' require understanding objects and actions, but do not require reasoning about different video moments.
In contrast, queries like ``the little girl talks after bending down'' require reasoning about the \emph{temporal relationship} between different actions (``talk'' and ``bend down'').
Localizing natural language queries in video is an important challenge, recently studied in \citet{hendricks2017localizing} and \citet{gao2017tall} with applications in areas such as video search and retrieval.
We argue that to properly localize queries with temporal language, models must understand and reason about intra-video context.

Reasoning about intra-video context is difficult as we do not know a priori which moments should be involved in the contextual reasoning and different queries may require reasoning about different contextual moments.
For example, in ``the little girl talks after bending down'', the relevant contextual moment ``bending down'' occurs just before the target moment ``the little girl talks''.
This is in contrast to the query ``the little girl talks {\it before} bending down'' where the relevant contextual moment occurs just after.
A limitation of current moment-localization models~\cite{hendricks2017localizing,gao2017tall} is they consider query-independent video context when localizing moments.
For example, when determining whether a proposed temporal region matches a natural language query, \citet{gao2017tall} considers the proposed temporal region, as well as video regions just before and after the proposed region.
Similarly, \citet{hendricks2017localizing} considers video context in the form of a global-context feature which represents the entire video.
While both may implicitly include the appropriate contextual moment in their context feature, they do not explicitly determine the relevant context for the query.

To address this difficulty, we propose Moment Localization with Latent Context (\modelname{}) which models video context as a \emph{latent variable}. %
The latent variable enables the model to attend to different video contexts conditioned on the specific query/video pair, offering flexibility in the location and length of the contextual moment and overcoming the limitation of query-independent contextual reasoning. 
We validate the importance of latent context by showing that our model performs well both on simple queries without temporal words and more complex queries requiring temporal reasoning. %
Moreover, our formulation is generic and unifies approaches in \citet{hendricks2017localizing} and ~\citet{gao2017tall}, allowing us to ablate model component choices, as well as which kind of video context is best for localizing moments described with temporal language.

Though datasets used for moment localization in video \cite{hendricks2017localizing,regneri2013grounding,sigurdsson2016hollywood} include temporal language, as we will show, there is not enough temporal language to effectively train and evaluate models.
We seek to extensively study this aspect, particularly with respect to temporal prepositions~\cite{pratt2004temporal}.
Thus, we collect the TEMPOral reasoning in video and language (\datasetname{}) dataset which builds off the recently collected DiDeMo dataset \cite{hendricks2017localizing}.  The dataset consists of two parts: a dataset with real videos and sentences created with a template model (\datasetname{} - Template Language (TL)), and a dataset with real videos and newly collected user-provided temporal annotations (\datasetname{} - Human Language (HL)).
Considering template sentences allows us to create a large dataset of sentences quickly for study of temporal language in a controlled setting.
The human language data then allows us to see these trends transfer to more complex human-language queries.
For data collection, we focus on the most common temporal referring words naturally occurring in language-and-video datasets.
Our contributions are twofold.
(i) We are the first to study models for temporal language in video moment retrieval with natural language queries. 
To this end, we introduce ~\datasetname{} which includes examples of how humans use temporal language to refer to video moments.
(ii) We propose ~\modelname{} for moment localization which treats video context as a latent variable and unifies prior approaches for moment localization. 
Our model outperforms prior work on~\datasetname{}-TL and ~\datasetname{}-HL as well as the original DiDeMo dataset.

%% file: related.tex
\section{Related Work}

\myparagraph{Localizing Video Segments with Natural Language.}  
Prior work has considered aligning natural language with video, e.g., instructional videos with transcribed text~\cite{kiddon15,huang17,malmaud14,malmaud15}.
Our work is most related to recent work in video moment retrieval with natural language \cite{gao2017tall,hendricks2017localizing}.
Both works take a natural language query and candidate video segment as input, and output a score for how well the natural language phrase aligns with the video segment.
\citet{gao2017tall} includes an additional loss to regress to start and end-points, whereas \citet{hendricks2017localizing} simplifies the problem by choosing from a discrete set of video segments.
Importantly, to represent a proposed video segment, both models consider context features around a moment: \citet{hendricks2017localizing} uses global context by averaging features over an entire input video, and \citet{gao2017tall} incorporates features adjacent to the proposed video segment. 
We argue that to do proper temporal reasoning, pre-determined, query independent context features may not cover all possible temporal relations.
Thus, we propose to model the context as a \emph{latent variable}, allowing our method to learn which context moments to consider as a function of the video and importantly, the query.

Both \citet{gao2017tall} and~\citet{hendricks2017localizing} collect data to test their models;
\citet{gao2017tall} considers the Charades~\cite{sigurdsson2016hollywood} and TACoS~\cite{regneri2013grounding} datasets. 
While TACoS includes localized sentences, Charades only has sentences and activity detection localizations, so a semi-automatic method is used to align action detection annotations to visual descriptions in Charades.
\citet{hendricks2017localizing} collected the Distinct Describable Moment (DiDeMo) dataset, which consists of Flickr~\cite{thomee2016yfcc100m} videos with localized referring expressions.
Both Charades and DiDeMo contain a large set of diverse videos (approximately 10,000 videos each).
We chose to base \datasetname{} on DiDeMo because it contains more clip/sentence pairs (40,000 vs. 13,000), and is focused on general videos which we believe is an interesting and useful scenario, rather than being restricted to indoor activities.

\myparagraph{Temporal Language.}  Prior work on temporal language processing has considered building explicit logical frameworks to process temporal prepositions like ``during'' or ``until'' (\citet{pratt2004temporal}, ~\citet{konur2008interval}).
We do not derive a particular temporal logic, but rather learn to understand temporal language in a data driven fashion. 
Furthermore, we specifically consider how to understand temporal words commonly used when referring to video content.
Other work has modeled dynamics for words which represent a change of state (e.g., ``pick up'') (~\citet{siskind2001grounding}, ~\citet{yu2015compositional}) in limited environments.
Though we limit the selection of temporal words in our study, the natural language in our data is open-world describing diverse events and how they relate to each other in video.
Interpretation of temporal expressions in text (``The game happened on the 19$^{th}$'') is a widely studied task (\citet{angeli2012parsing}, \citet{zhong2017time}).
Our work is distinctly different from this line of work as we specifically study temporal prepositions and how they refer to video.

\myparagraph{Modeling Visual Relationships.}  A variety of papers have considered modeling spatial relationships in natural images~\cite{dai2017detecting,hu2016modeling,peyre2017weakly,plummer2017phrase}.
Our approach is analogous to this in the temporal domain; we hope to localize moments in videos.
CLEVR, a synthetic visual question answering (VQA) dataset~\cite{johnson2016clevr}, was created to allow researchers to systematically study the ability of models to perform complex reasoning.
Our dataset is partially motivated by the success of CLEVR to enable researchers to study reasoning abilities of different models in a controlled setting.
In contrast to CLEVR we consider a more diverse visual input in the form of real videos.

In the video domain, the TGIF-QA~\cite{jang2017tgif} and Mario-QA~\cite{mun2016marioqa} datasets provide opportunities to study temporal reasoning for the task of VQA.
The TGIF-QA dataset considers three types of temporal questions: before/after questions, repetition count, and determining a repeating action.
Each question is accompanied by multiple choice answers.
Videos we consider are much longer (25-30s as opposed to an average of 3.1s) which makes the use of temporal reasoning much more important.
The MarioQA dataset is an additional VQA dataset designed to gauge temporal reasoning of VQA systems.
Both TGIF-QA and MarioQA datasets include template-based natural language queries.
In this paper, we consider synthetic queries similar to TGIF-QA and MarioQA, but also include human language queries. %
In addition, unlike the MarioQA dataset, that consists of synthetic data constructed from gameplay videos, our dataset consists of real visual inputs, and includes temporal grounding of natural language phrases.
Finally, neither TGIF-QA nor MarioQA include temporal localization.

%% file: method.tex
\newcommand{\modelparams}{\phi}
\newcommand{\paramsmodel}{\modelparams}
\newcommand{\paramssimilarity}{\paramsmodel_{\mathcal{S}}}
\newcommand{\paramsvideo}{\paramsmodel_{\mathcal{V}}}
\newcommand{\paramslanguage}{\paramsmodel_{\mathcal{L}}}
\newcommand{\featurevideo}{f_{\mathcal{V}}}
\newcommand{\featurelanguage}{f_{\mathcal{L}}}
\newcommand{\featurevisual}{g}
\newcommand{\similaritybase}{s}
\newcommand{\similaritybasecontext}{f_{\mathcal{S}}}
\newcommand{\featureendpoint}{f_{\mathcal{T}}}
\newcommand{\distbase}{d} %
\newcommand{\distbasecontext}{\widetilde{d}} %
\newcommand{\basemoment}{\tau}
\newcommand{\contextmoment}{\tau^\prime}
\newcommand{\contextset}{\Tau_{\basemoment}}
\newcommand{\video}{v}
\newcommand{\sentence}{q}
\newcommand{\basemomentstart}{\basemoment^{(s)}}
\newcommand{\basemomentend}{\basemoment^{(e)}}
\newcommand{\contextmomentstart}{\tau^{\prime(s)}}
\newcommand{\contextmomentend}{\tau^{\prime(e)}}

\section{Moment Localization with Latent Context}

Given a video $\video$ and natural-language query $\sentence$ describing a moment in the video, our goal is to output the moment $\basemoment=\left(\basemomentstart,\basemomentend\right)$ where $\basemomentstart$ and $\basemomentend$ are temporal start and end points in the video, respectively. 
In the following, we formulate a generic, unified model which encompasses prior approaches~\cite{hendricks2017localizing,gao2017tall}.
This allows us to explore and evaluate trade offs for different model components and extensions which then leads to higher performance.
Unlike prior work, we consider a latent context variable which enables our model to better reason about temporal language.

\begin{figure}[t]
\centering
\includegraphics[width=1.0\linewidth]{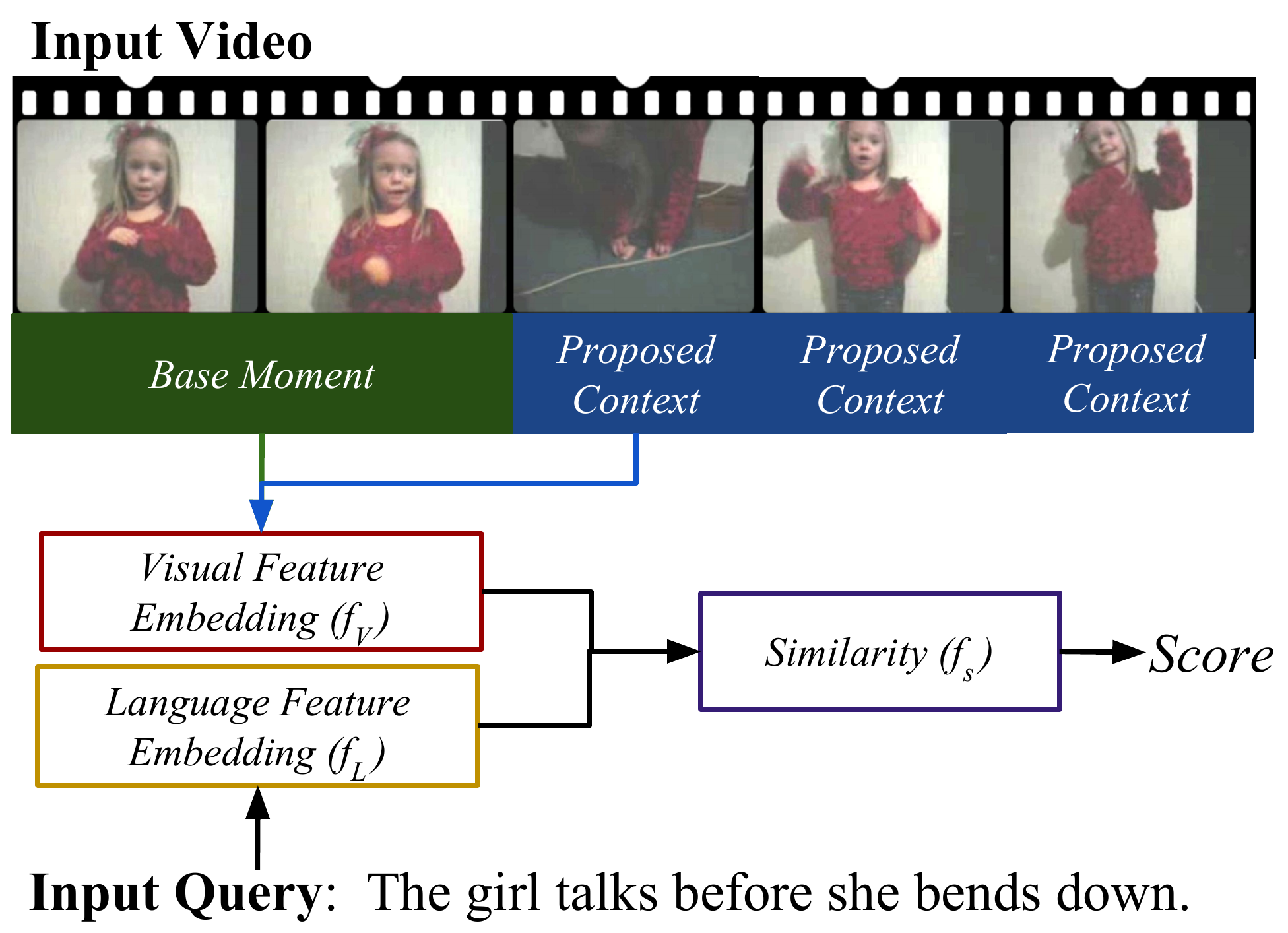}
\caption{Our model, Moment Localization with Latent Context (MLLC), takes a video and a text query as input and outputs the moment in the video corresponding to the query.
\modelname{} considers many different \textit{context} moments (blue) for a specific \textit{base} moment (green).
}  

\label{fig:model}
\end{figure}

Let the moment $\basemoment$ corresponding to the text query be the \textit{base} moment and the set of other video moments $\contextset$ be possible \textit{context} moments for $\basemoment$.
We define a scoring function between the video moment and natural-language query by maximizing over all possible context moments $\contextmoment\in\contextset$,
\begin{equation}
 \similaritybase_{\paramsmodel}\left(\video,\sentence,\basemoment\right) = \max_{\contextmoment \in \contextset} \similaritybasecontext\left(\featurevideo\left(\video,\basemoment, \contextmoment\right),\featurelanguage\left(\sentence\right)\right),
\label{eqn:latent}
\end{equation}
where $\featurevideo$ and $\featurelanguage$ are functions computing features over the video and language query, $\similaritybasecontext$ is a similarity function, and $\paramsmodel$ are model parameters.
This formulation is generic and trivially encompasses the MCN and TALL formulations by letting the set of possible context moments $\contextset$ be their respective single-context moment.
Figure~\ref{fig:model} shows the generic structure of our model. %

With this formulation, we seek to answer the following questions: (i) Which combination of model components performs best for the moment-retrieval task? Though our primary goal is localizing moments with temporal language, we believe a good base moment retrieval model is important for localizing moments with temporal language.  (ii) How best to incorporate context for moment retrieval with temporal language? 
We first detail the different terms and outline different model design choices, where design choices marked with \textbf{\textit{bold-italic font}} is ablated in Section~\ref{sec:evaluation}. 
Components which are used in our final proposed Moment Localization with Latent Context (MLLC) model and prior models are summarized in Table~\ref{tab:models}.

\myparagraph{Video feature $\featurevideo$.} 
The video feature $\featurevideo=\left(\featurevisual\left(\video, \basemoment\right),\featurevisual\left(\video,\contextmoment\right),\featureendpoint\left(\basemoment,\contextmoment\right)\right)$ is a concatenation of visual features for the base $\featurevisual\left(\video, \basemoment\right)$ and context $\featurevisual\left(\video,\contextmoment\right)$ moments and endpoint features $\featureendpoint\left(\basemoment,\contextmoment\right)$.
To compute visual features $\featurevisual$ for a temporal region $\basemoment$, per-frame features are averaged over the temporal region.
Note that if the context moment consists of more than one contiguous temporal region, then the visual features are computed over each contiguous temporal region and then concatenated (c.f., before/after context in TALL, explained below). 
There are many choices for visual features. 
TALL~\cite{gao2017tall} compares average $fc_7$ features (extracted from \cite{simonyan2014very}) to features extracted with C3D~\cite{tran2015learning} and LSTM features~\cite{donahue2015long}.
Surprisingly, C3D features only outperform average $fc_7$ features by a small margin.
We use the visual features used in the MCN model \cite{hendricks2017localizing}, which are similar to the $fc_7$ features from~\cite{gao2017tall}, but included motion features as well, computed from optical flow (extracted with \cite{wang2016temporal}).
We then pass the extracted visual features through a MLP.
Note that we learn separate embedding functions for RGB and optical flow inputs and combine scores from different input modalities using a late-fusion approach~\cite{hendricks2017localizing}.

\myparagraph{Endpoint feature $\featureendpoint$.} 
Modeling temporal context requires understanding how different temporal segments relate in time.
\citet{hendricks2017localizing} suggest including temporal endpoint features (\textbf{\textit{TEF}}) $\featureendpoint = \left(\basemomentstart,\basemomentend\right)$ for the base moment which encode when the moment starts and ends to better localize sentences which include words like ``first'' and ``last''. 
Note that TALL~\cite{gao2017tall} does not incorporate TEFs.
In order to understand temporal relationships, it is important that models also include features which indicate when a context moment occurs.
In addition to providing TEFs for base moments, we also experiment with concatenating TEFs for context moments (\textbf{\textit{conTEF}}) $\featureendpoint = \left(\basemomentstart,\basemomentend,\contextmomentstart,\contextmomentend\right)$.

\myparagraph{Language feature $\featurelanguage$.} Text queries are transformed into a fixed-length vector with an LSTM~\cite{hochreiter1997long}.
Before inputting words into the LSTM, they are embedded in the Glove~\cite{pennington2014glove} embedding space. 
The final layer of the LSTM is projected into the shared video-language embedding space with a fully connected layer. 
\citet{gao2017tall} considers LSTM language features and Skip-thought encoders.
Our main goal is to study how context impacts moment localization with temporal language, so we use the LSTM features used on the original DiDeMo dataset. %

\myparagraph{Similarity $\similaritybasecontext$.} 
Given video $\featurevideo$ and language $\featurelanguage$ features, we consider three ways to encode similarity between the features. 
Like \citet{hendricks2017localizing}, we consider a \textbf{\textit{distance-based}} similarity $\similaritybasecontext = \left(|\featurevideo-\featurelanguage|^2\right)$. 
Second, we consider a fused-feature similarity (\textbf{\textit{mult}}) where the Hadamard product $\featurevideo\odot\featurelanguage$ between the two features are passed to a MLP. 
We also explore unit normalizing features before the Hadamard product (\textbf{\textit{normalized mult}}).
Finally, we consider the similarity (\textbf{\textit{TALL similarity}}) which consists of the concatenation $\left(\featurevideo,\featurelanguage,\featurevideo\odot\featurelanguage,\featurevideo+\featurelanguage\right)$ and then passed to a MLP.

\myparagraph{Context moments $\contextset$.} 
We consider three sets of context moments. 
First, we consider the entire video as the context moment (\textbf{\textit{global}}) following ~\citet{hendricks2017localizing}. 
Second, we consider using the moments just before and after the base moment (\textbf{\textit{before/after}}). 
Finally, we consider using the set of all possible moments (\textbf{\textit{latent}} context) which offers greatest flexibility in contextual reasoning.

\myparagraph{Training loss.} 
We consider two training losses. 
The first loss is the MCN \textbf{\textit{ranking loss}} 
which encourages positive moment/query pairs to have a smaller distance in a shared embedding space than negative moment/query pairs.
To sample negative moment/sentence pairs, they consider negative moments \textit{within} a specific video (called intra-video negative moments) and negative moments in different videos (called inter-video negative moments).
This sampling strategy leads to a small improvement in performance (approximately one point on all metrics) when compared to just using intra-video negative moments.
We also consider the alignment loss used in TALL (\textbf{\textit{TALL loss}}) which is the sum of two log-logistic functions over positive and negative training query/moment pairs (intra-video negatives are used). 

\myparagraph{Supervising context moments.} 
For the temporal sentences in our newly collected dataset (Section~\ref{sec:dataset}), we have access to the ground-truth context moment during training. 
Thus, we can contrast a \textbf{\textit{weakly supervised}} setting in which we optimize over the unknown latent context moments during learning and inference to a \textbf{\textit{strongly supervised setting}}.

\myparagraph{Implementation details.} 
Candidate base and context moments coincide to the pre-segmented five-second segments used when annotating DiDeMo. 
Moments may consist of any contiguous set of five-second segments.
For a 30-second video partitioned into six five-second segments, there are 21 possible moments.
All models were implemented in Caffe~\cite{jia2014caffe} and optimized with SGD. 
Models were trained for $\sim$ 90 epochs with an initial learning rate of 0.05, which decreases every 30 epochs.
Code is publicly released\footnote{\url{https://people.eecs.berkeley.edu/~lisa_anne/tempo.html}}.

\begin{table*}[t]
\centering %
\resizebox{\linewidth}{!}{%
\begin{tabular}{l c c c c c} %
\toprule %
& Endpoint &  Similarity & Context & Training & Supervised \\
& Feature &   &  & Loss & Temp.\ Context\\
\hline
TALL~\cite{gao2017tall} & None & TALL sim. & Before/After & TALL loss & None \\
MCN~\cite{hendricks2017localizing} & TEF & Distance-based & Global & Ranking & None \\
MLLC (ours) & \textbf{conTEF} & Normalized mult & \textbf{Latent} & Ranking & \textbf{Strongly sup.} \\
\bottomrule %
\end{tabular}
}
\vspace{.1cm}
\label{tab:models}
\caption{Comparison of models.  Bolded entries show our additions for localizing temporal language.} %
\end{table*}

%% file: dataset.tex
\section{The \datasetname{} Dataset}
\label{sec:dataset}

We collect the TEMPOral reasoning in video and language (\datasetname{}) dataset based off the recently released DiDeMo dataset.
Our dataset consists of two parts: \datasetname{} - Template Language (TL) and \datasetname{} - Human Language (HL).
We create \datasetname{} - TL using language templates to augment the original sentences in DiDeMo with temporal words.
The template allows us to generate a large number of sentences with known ground truth base and context moments.
However, template language lacks the complexity of human language, so we then collect an additional fully user-constructed dataset, \datasetname{} - HL, consisting of sentences that contain specific temporal words. %

\textbf{Temporal Words in Current Datasets.} We first analyze temporal words which occur in current natural language moment retrieval datasets.
We consider temporal adjectives, adverbs, and prepositions found both by closely analyzing moment-localization datasets and consulting lists containing words which belong to different parts of speech.  
In particular, we rely on the preposition project~\citep{litkowski2005preposition}\footnote{\url{http://www.clres.com/prepositions.html}} to scrape relevant temporal words.
Table~\ref{tab:word_count} shows example temporal words and the number of times they occur in each dataset (TACoS~\cite{regneri2013grounding}, Charades~\cite{gao2017tall}, DiDeMo~\cite{hendricks2017localizing}). 
Though all moment localization datasets use temporal words, they do not contain enough examples to reliably train and evaluate current models.
Additionally, we observe that temporal words which are frequently used when describing video segments are different than those commonly used in text without video grounding.
For example, in~\citet{pratt2004temporal}, ``during'' is a common example, but we observe that ``during'' is infrequently used when describing video.
Of temporal words, we focus on the four most common words, ``before'', ``after'', ``then'', and ``while'' when creating our dataset.

\begin{table}[t]
\centering %
\scalebox{0.65}{
\begin{tabular}{c rrrrrrr} %
\toprule %
Dataset& Before & After & Then & While & Yet & During & Until\\ 
\midrule %
TACoS %
& 50 & 62 & 731 & 82 & 23 & 0 & 4\\ %
Charades %
& 281 & 27 & 1873 & 1165 & 0 & 3 & 1\\
DiDeMo %
& 198 & 119 & 1021 & 266 &16 & 21 & 22\\ %
\datasetname{} - TL & 23,842 & 23842 & 11921 & - & - & - & -\\
\datasetname{} - HL & 6610 & 5495 & 5478 & 5425 & - & - & -\\[1ex] %
\bottomrule %
\end{tabular}
}
\vspace{.1cm}
\caption{\label{tab:word_count}Word frequency of temporal words in natural language moment localization datasets.} %
\end{table}

\textbf{\datasetname{} - Template Language.}  To construct sentences in \datasetname{}-TL, we find adjacent moments in the DiDeMo dataset and fill in template sentences for ``before'', ``after'', and ``then'' temporal words.
For ``before'', we use two templates: ``$X$ before $Y$'' and ``Before $Y$, $X$'', where $X$ and $Y$ are sentences from the original DiDeMo dataset.
Likewise for ``after'', we consider the templates ``$X$ after $Y$'' and ``After $Y$, $X$''.
For ``then'' we only consider one template, ``$X$ then $Y$.''

\begin{figure}[t]
\centering
\includegraphics[width=\linewidth]{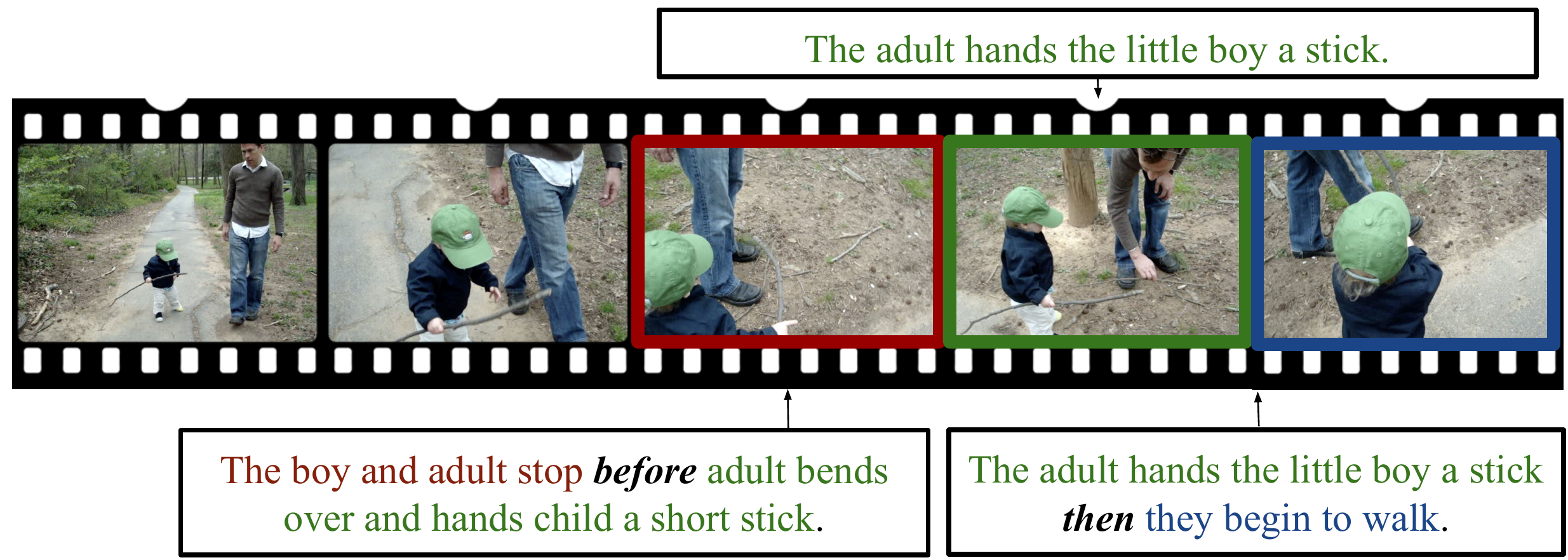}
\caption{Example sentences in \datasetname \ - HL.  The top sentence corresponds to the reference moment (shown in green).  The bottom sentences are newly collected sentences which use temporal language.%
}
\label{fig:annotations}
\end{figure}

\textbf{\datasetname{} - Human Language.}  Though the template dataset is an interesting testbed for understanding temporal language, it is difficult to replicate the interesting complexities in human language.
For example, when writing long sentences with temporal prepositions, humans frequently make use of language structure such as coreference to form more cohesive statements.
To collect annotations, we follow the protocol in~\citet{hendricks2017localizing} and segment videos into 5-second temporal segments.
After collecting descriptions, we ensure descriptions are localizable by asking other workers to localize each moment.
To collect data for ``before'', ``after'', and ``then'', we ask annotators to describe a segment in \textit{relation} to a ``reference'' moment from the DiDeMo dataset.
For example, if the DiDeMo dataset includes a localized phrase like ``the cat jumps'', annotators write a sentence which refers to the segment ``the cat jumps'' using a specific temporal word.
We provide both the phrase (``the cat jumps'') and the reference moment to annotators, and the annotators provide a sentence describing a new moment which references the reference moment.

\datasetname{}-HL includes unique properties which are hard to replicate with template data.
Figure~\ref{fig:annotations} depicts the base moment provided to workers, as well as descriptions from \datasetname{}-HL.
In Figure~\ref{fig:annotations}, the description ``The adult hands the little boy the stick then they walk away'' includes an example of visual coreference (``they'').
We note that use of pronouns is much more prevalent in \datasetname{}-HL, with 28.1\% of sentences in \datasetname{}-HL including pronouns (``he'', ``she'', ``it'') in contrast to 10.3\% of sentences in the original DiDeMo dataset.
Additionally, annotators will refer to the base moment with different language than originally used in the base moment (e.g., ``the girl waves at the camera'' versus the base moment ``the girl looks at the camera and waves'') in order to make their sentences more fluent.

%% file: results.tex
\section{Experiments}
\label{sec:evaluation}

\myparagraph{Evaluation Method.}
We follow the evaluation protocol defined for the DiDeMo dataset~\cite{hendricks2017localizing} over all possible combinations of the five-second video segments. 
We report rank at one (R@1), rank at five (R@5), and mean intersection over union (mIOU) using their aggregator over three out of the four human annotators. 
We compare our models on \datasetname{}-TL, \datasetname{}-HL, and the DiDeMo dataset.
When training our models, we combine the DiDeMo dataset with \datasetname{}-TL or \datasetname{}-HL.
This enables our model to concurrently learn to localize the simpler DiDeMo sentences with more complex \datasetname{} sentences.

\myparagraph{Baselines.} We compare to the two recently proposed approaches for video moment localization: MCN \cite{hendricks2017localizing} and TALL \cite{gao2017tall}.
We adapt the implementation of TALL \cite{gao2017tall} to the DiDeMo dataset in three ways.  
First, we do not include the temporal localization loss required to regress to specific start and end points as DiDeMo, and thus also TEMPO, is pre-segmented, so the model does not need to compute exact start and end points. %
Second, the original TALL model uses C3D features.
For a fair comparison we train both models with the same RGB and flow features extracted as was done for the original MCN model.
Finally, the MCN model proposes temporal endpoint features (TEF) to indicate when a proposed moment occurs within a video.
We train TALL with and without the TEF and show that TEF improves performance on the original DiDeMo dataset.

\myparagraph{Ablations.}  To ablate our proposed latent context, we compare to other models which share the same MLLC base network.  
We consider the MLLC model with global context and before/after context.
We also train a model with weakly supervised (WS) latent context and strongly supervised (SS) latent context.
We also train models both with and without context TEF (conTEF).

\myparagraph{The MLLC Base Model.}  We first ablate our MLLC base model (Table ~\ref{tab:base_ablate}).
We train our models on \datasetname{}-TL and DiDeMo and evaluate on the original DiDeMo dataset.
All models are trained with global context.
We find that the ranking loss is preferable on the DiDeMo dataset (compare lines 1 and 2) and that TALL-similarity performs better than the distance based similarity of the MCN model (compare lines 1 and 5).
A simpler version of the TALL-similarity, in which the concatenated element wise multiplication, element wise sum, and concatenation is replaced by a single normalized elementwise multiplication, increases R@1 by almost one point and increases mIoU by over two points (compare lines 5-7).
We call our best model the MLLC-Base model (line 7).
Our MLLC-Base model performs better than previous models (MCN line 1 and TALL line 3).

\begin{table}[h]
\centering %
\scalebox{0.7}{
\begin{tabular}{l l l l| rrr} %
\toprule
& Model & Similarity & Training & R@1 & R@5 & mIoU \\ 
&  &  & Loss & & \\ \hline
1 & MCN &  Dist.-based & Ranking & 26.63 & 73.38 & 41.14 \\
2 & MCN &  Dist.-based & TALL & 23.89 & 76.54 & 35.69 \\
3 & TALL & TALL-sim. & TALL & 8.04 & 36.32 & 22.68\\ 
4 & TALL w/TEF  & TALL-sim. & TALL & 23.56  & 72.74 & 35.58\\ 
5 & MCN  & TALL-sim & Ranking & 27.52 & \textbf{79.07} & 41.48\\
6 & MCN  & Mult  & Ranking & 28.19 & 78.97 & 43.21 \\
7 & MLLC-Base  & Norm. Mult & Ranking & \textbf{28.37} & 78.64 & \textbf{43.65}\\

\bottomrule
\end{tabular}
}
\vspace{.1cm}
\caption{\label{tab:base_ablate} To select our base network, we consider different variants on the two previously proposed moment retrieval methods, TALL~\cite{gao2017tall} and MCN~\cite{hendricks2017localizing}.  Results reported on val.} %
\end{table}

\begin{table*}[t]
\centering %
\resizebox{\linewidth}{!}{%
\begin{tabular}{l l | rr | rr | rr | rr || rrr} %
\toprule %
& & \multicolumn{2}{c}{} &
\multicolumn{6}{c}{\datasetname{} - Template Language (TL)} \\ 

& & \multicolumn{2}{c}{DiDeMo} &
\multicolumn{2}{c}{Before} &
\multicolumn{2}{c}{After} & 
\multicolumn{2}{c}{Then} &
\multicolumn{3}{c}{\textbf{Average}} \\ 
& & R@1 & mIoU &
R@1 & mIoU & 
R@1 &  mIoU &
R@1 &  mIoU &
R@1 & R@5 & mIoU\\
\midrule

 1 & Frequency Prior & 
  10.71 & 20.67  & 
   17.85 & 24.22 &
 22.42& 25.76&
 0.00 & 24.73 & 
 12.74& 52.58 & 23.84  \\

 2 & MCN  & 
  24.85 & 37.92  & 
   32.28 & 38.67 &
 26.08 & 35.44 &
  25.07 & 53.94 & 
 27.07& 73.36 & 41.49  \\ 

3 & TALL  & 
 20.95 &  32.09  &
  27.13 & 32.41  &
 26.30 & 34.27  &
 4.84& 36.75 &
 19.80 & 64.66& 33.88\\ 

4 & \modelname{}- Global & 
  26.32 & 40.37  &
  31.92 & 38.26 &
 25.37 &  35.59 &
 \textbf{27.53} & \textbf{57.08} &
 27.78 & 74.14  & 42.82 \\
 
 5 & \modelname{} B/A & 
  26.04 & 39.60  &
  34.04 &  40.46 &
 28.50&  38.18 &
 25.60& 54.37 &
 28.54& 74.92  & 43.15  \\
 
 6 & \modelname{} (WS) & 
  26.57 & 40.99  &
 30.56 & 37.64 &
 24.76& 35.10  &
 26.95&  56.49&
26.95 & 74.18  & 42.55 \\

 7 & \modelname{} (WS + conTEF) & 
  25.87 & 40.37  &
  32.01 & 39.51 &
 24.31& 33.94  &
 24.98& 55.22 &
26.79 & 74.04 & 42.27 \\
 
 8 & \modelname{} (SS) & 
 26.09  & 40.12  &
  28.45& 34.38 &
 23.79& 33.92  &
 24.27 & 55.00 &
25.65 & 73.60   & 40.86 \\
 
9 & \modelname{} (SS + conTEF) & 
 \textbf{27.46} & \textbf{41.20} &
 \textbf{35.31} & \textbf{41.81}  & 
 \textbf{29.38} & \textbf{38.90} &
 26.83& 54.97  &
\textbf{29.74} & \textbf{76.76}  & \textbf{44.22} \\

\bottomrule 
\end{tabular}
}
\vspace{.1cm}
\caption{\label{tab:results_toy}Comparison of different model performance for different temporal words on \datasetname \ - TL on our test set.  We report scores for the three temporal words in \datasetname{} - TL as well as on the original DiDeMo dataset.  We find that our model performs best when considering all sentence types. B/A indicated before/after context, WS indicates weak context supervision, and SS indicates strong context supervision.} %
\end{table*}

\myparagraph{Results: \datasetname{} - TL.}  We first compare different moment localization models on \datasetname{} - TL (Table~\ref{tab:results_toy}).
In particular, our model performs well on ``before'' and ``after'' words.
Additionally, our MLLC model with global context outperforms both the MCN model~\cite{hendricks2017localizing} and the TALL~\cite{gao2017tall} model when considering all sentence types, verifying the strength of our base \modelname{} model.

Comparing MLLC with global context and MLLC with before/after context (compare row 4 and 5), we note that before/after context is important for localizing ``before'' and ``after'' moments.
However, our model with strong supervision (row 9) outperforms the model trained with before and after context, suggesting that learning to reason about which context moment is correct (as opposed to being explicitly provided with the context before and after the moment) is beneficial.
We note that strong supervision (SS) outperforms weak supervision (WS) (compare rows 7 and 9) and that the context TEF is important for best performance (compare rows 8 and 9).

We note that though the MLLC-global model outperforms our full model for ``then'' on \datasetname{}-TL, our full model performs better on then for the \datasetname{}-HL (Table~\ref{tab:results_real}).
One possibility is that the ``then'' moments in \datasetname{}-TL do not require context to properly localize the moment.
Because \datasetname{}-TL is constructed from DiDeMo sentences, constituent sentence parts are \textit{referring}.  
For example, given an example sentence from \datasetname{}-TL (e.g., ``The cross is seen for the first time \textit{then} window is first seen in room''), the model does not need to reason about the ordering of ``cross seen for the first time'' and ``window is seen for the first time'' because both moments only happen once in the video.
In contrast, when considering the sentence ``The adult hands the little boy a stick \textit{then} they begin to walk'' (from Figure~\ref{fig:annotations}), ``begin to walk'' could refer to multiple video moments.
Consequently, our model must reason about the temporal ordering of reference moments to properly localize the video moment.

\begin{table}[t]
\centering %
\scalebox{0.7}{
\begin{tabular}{lrr|rr|rr} %
\toprule
& \multicolumn{2}{c}{Before} &
\multicolumn{2}{c}{After} &
\multicolumn{2}{c}{Then} \\ 

Context & R@1 & mIoU & 
R@1 & mIoU & 
R@1 & mIoU\\

\midrule

Global & 
 -1.07	& -2.72 & 	-7.59	& -6.75 & 	43.30	& 31.57  \\

 Before/After & 
2.77 &	2.03	& 11.47	& 12.08	& 42.92	& 29.09 \\

 Latent & 
 7.78 &	37.55	& 8.58 &	10.39 &	50.09	& 33.64\\

\bottomrule
\end{tabular}
}
\vspace{.1cm}
\caption{\label{tab:results_prior} Difference between performance on full dataset and set on which reference moments are localized properly for different methods on \datasetname{}-TL.
} %
\end{table}

\begin{table*}[t]
\centering %
\resizebox{\linewidth}{!}{%
\begin{tabular}{p{2.8cm} rr | rr | rr | rr | rr | rrr} %
\toprule
& \multicolumn{2}{c}{} &
\multicolumn{8}{c}{\datasetname{} - Human Language (HL)} \\ 
\multicolumn{4}{c}{} \\ 

& 
\multicolumn{2}{c}{DiDeMo} &
\multicolumn{2}{c}{Before} &
\multicolumn{2}{c}{After} &
\multicolumn{2}{c}{Then} &
\multicolumn{2}{c}{While} &
\multicolumn{3}{c}{\textbf{Average}} \\ 

& R@1  & mIoU & 
R@1 & mIoU & 
R@1 & mIoU & 
R@1 & mIoU & 
R@1 & mIoU & 
R@1 & R@5 & mIoU\\

\midrule

Frequeny Prior &
 19.43& 25.44  &
  29.31& 51.92 & 
 0.00& 0.00 &
 0.00 &  7.84 &
 4.74 & 12.27 &
 10.69& 37.56 & 19.50 \\ 

MCN & 
  26.07  & 39.92 & 
 26.79 & 51.40 & 
 \textbf{14.93} & 34.28 & 
 18.55 & 47.92 & 
 10.70 & 35.47 &
 19.4 & 70.88 & 41.80  \\ 

TALL  + TEF & 
21.79 & 33.55 & 
 25.91&  49.26&
 14.43 & 32.62 &
 2.52& 31.13 &
 8.1 & 28.14 &
 14.55 & 60.69 & 34.94 \\ 

\modelname{} - Global  & 
27.01 & 41.72 &
27.42 & 52.22 &
 14.10 & 34.33  &
18.40 & 49.17 &
 10.86& 35.36 &
 19.56 & 71.23 & 42.56 \\ 

\modelname{} - B/A & 
26.47  & 40.39 &
 31.95 & 55.89  &
 \textbf{14.93} & 34.78 &
17.36 & 47.52 &
 \textbf{11.32} & 35.52 &
 20.40 & 70.97 & 42.82 \\

\modelname{} (Ours) & 
 \textbf{27.38} & \textbf{42.45} & 
 \textbf{32.33} & \textbf{56.91} & 
 14.43 & \textbf{37.33} & 
 \textbf{19.58} & \textbf{50.39} & 
 10.39& \textbf{35.95} &
 \textbf{20.82} & \textbf{71.68} & \textbf{44.57} \\ 
 \midrule \midrule
 \modelname{} (Ours) Context Sup. Test & 
 27.39 & 42.25 & 
 52.58 & 80.37 & 
 36.48 & 75.79 & 
 36.05 & 70.51 & 
 10.39 & 35.87 &
 32.58 & 79.86 & 60.96 \\ 

\bottomrule
\end{tabular}
}
\vspace{.1cm}
\caption{\label{tab:results_real}Comparison of different model performance on \datasetname{} - HL on the test set.  ``MLLC - Global'' indicates our model with global context and ``MLLC - B/A'' indicated MLLC with before/after context.} %
\end{table*}

On \datasetname{} - TL, sentences differ from original DiDeMo sentences solely because of the use of temporal words.
Thus, we can do a controlled study of how well models understand temporal words.
If a model has good temporal reasoning, then if it can localize a reference moment ``the dog jumps'' it should be easier for the model to localize the moment ``the dog sits after the dog jumps''.
To test whether models are capable of this, we look at only sentences in \datasetname{} - TL where the model has correctly localized the corresponding context moment in DiDeMo (Table~\ref{tab:results_prior}). 
We report the \textit{difference} in performance when considering only sentences in which temporal context was properly localized and all sentences.
On our model, performance on all three temporal word types increases when the context moment can be properly localized.
When considering global context, performance on ``before'' and ``after'' actually decreases, suggesting global context does not understand temporal reasoning well. %
Finally, even when the context is correctly localized, there is still ample room for improvement on all three sentence types motivating future work on temporal reasoning for moment retrieval.

\myparagraph{Results: \datasetname{} - HL.}  Table~\ref{tab:results_real} compares performance on \datasetname{} - HL.  
We compare our best-performing model from training on the \datasetname{}-TL (strongly supervised MLLC and conTEF) to prior work (MCN and TALL) and to MLLC with global and before/after context.
Performance on \datasetname{}-HL is considerably lower than \datasetname{}-TL suggesting that \datasetname{}-HL is harder than \datasetname{}-TL.

On \datasetname{} - HL, we observe similar trends as on \datasetname{}-TL.
When considering all sentence types, \modelname{} has the best performance across all metrics.
In particular, our model has the strongest performance for all sentence types considering the mIoU metric.
In addition to performing better on temporal words, our model also performs better on the original DiDeMo dataset.
As was seen in \datasetname{}-TL, including before/after context performs better than our model trained with global context for both ``before'' and ``after'' words.

The final row of Table~\ref{tab:results_real} shows an upper bound in which the ground truth context is used at test time instead of the latent context. %
We note that results improve for ``before'', ``after'', and ``then'', suggesting that learning to better localize context will improve results for these sentence types.

\begin{table}[t]
\centering %
\small
\begin{tabular}{l|r r|r r} %
\toprule
& \multicolumn{2}{c}{Before} & \multicolumn{2}{c}{After} \\ 
& R@1 & mIoU & R@1  & mIoU \\
\midrule
Context Fragment & 25.16 & 32.94 & 23.05 & 27.64 \\
Full Sentence &  \textbf{27.55}& \textbf{35.70} & \textbf{32.67} &\textbf{40.39} \\
\bottomrule
\end{tabular}
\vspace{.1cm}
\caption{\label{tab:results_context}Comparison of different methods to localize context fragments (e.g., the text ``she bends down'' in the sentence ``the girl talks after she bends down'').  We compare localizing fragments with the MLLC model to localizing fragments with the latent context considered when localizing the whole query.} %
\end{table}

\textit{Localizing Context Fragments.} \datasetname{}-HL sentences can be broken into two parts: a base-sentence fragment (which refers to the base moment), and a context-sentence fragment (which refers to the context moment).
For example, for the sentence ``The girl holds the ball before throwing it,'', ``the girl holds the ball'' is the base fragment and ``throwing it'' is the context fragment.
A majority of the ``before'' and ``after'' sentences in \datasetname{}-HL are of the form ``$X$ before (or after) $Y$'', so we can determine a list of sentence fragments by splitting sentences based on the temporal word.
Given ``before'' and ``after'' sentences, we determine the ground truth context fragment by considering which reference moment was given to annotators.
We can then measure how well models localize context fragments. 
Table~\ref{tab:results_context} compares two approaches to localizing context fragments: inputting just the context fragment into MLLC and reporting the context used by MLLC when inputting the entire query into our model.
We find that our model reliably selects the correct context fragments, most likely because it can properly exploit temporal understanding of how the context fragment relates to the base fragment.

\begin{figure}[t]
\centering
\includegraphics[width=1.0\linewidth]{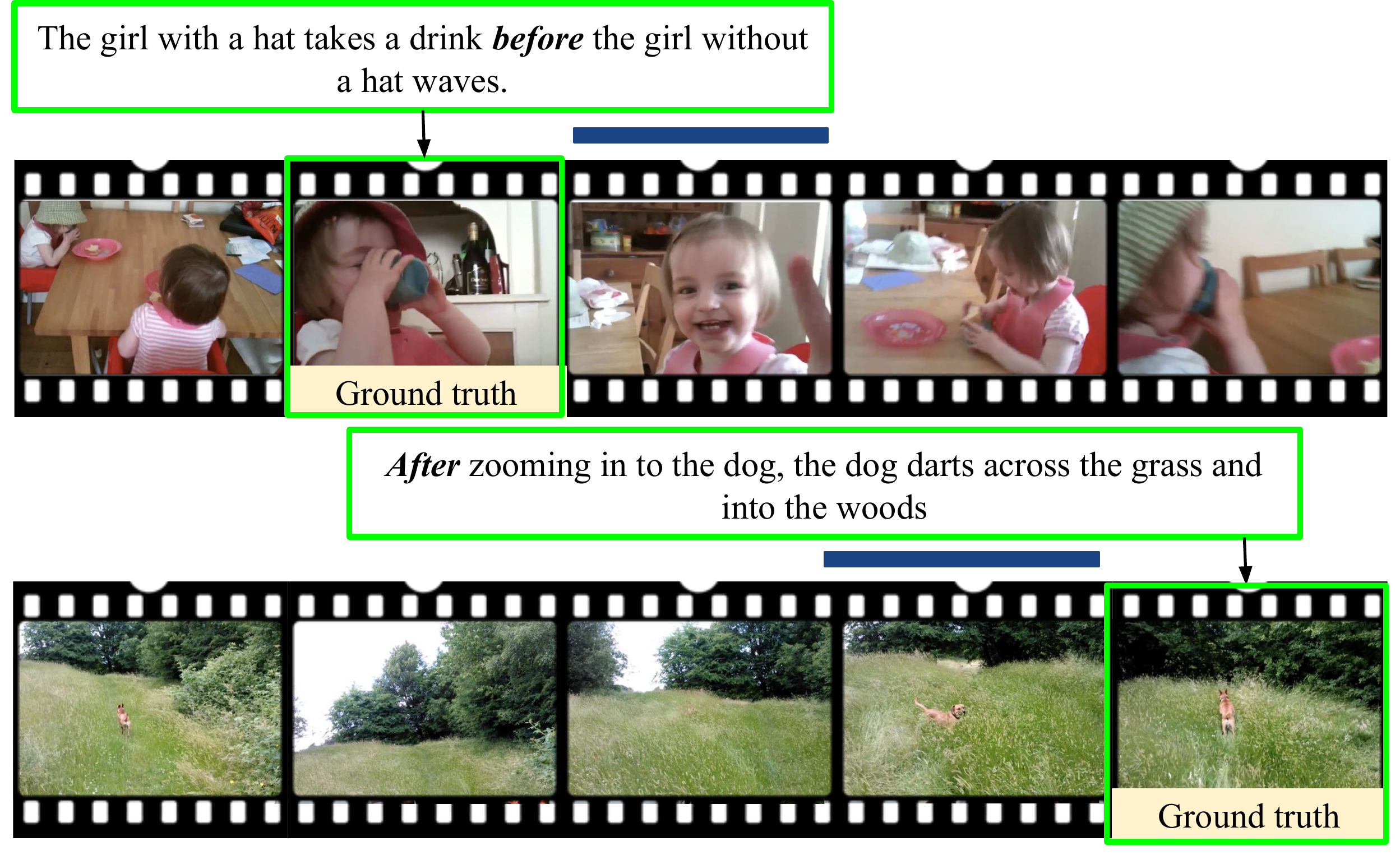}
\caption{Moment localization predictions on \datasetname{} - HL using our model.
In addition to the localized query, we show the selected context segment (blue line) that our model considers when localizing the query.
}
\label{fig:vis_context_real}
\vspace{-2mm}
\end{figure}

\textit{Visualizing Context.}  In addition to a localized query, we can also visualize which context moment the temporal query refers to. 
Figure~\ref{fig:vis_context_real} shows predicted moments and their corresponding context moments.
For the query ``The girl with a hat takes a drink before the girl without a hat waves'', the little girl in the hat drinks twice, but our model correctly localizes the time she drinks \textit{before} the other girl waves.
Likewise, for the moment ``After zooming in to the dog, the dog darts across the grass and into the woods'', the dog darts towards the woods twice (at the beginning of the video and at the end).
Our model properly localizes the moment when the dog runs towards the forest the second time as well as the context fragment ``zooming in on dog'' when localizing the moment.

\myparagraph{Discussion.}  We show promising results on both \datasetname{}-TL and \datasetname{}-HL, but there is potential improvement for building better frameworks for understanding temporal language.
In Table~\ref{tab:results_real}, strongly supervising context at test time improves overall results, suggesting that models which can better localize context text will outperform our current model.
Though \datasetname{} and DiDeMo have over 60,000 sentences combined, visual content is quite diverse.  
Integrating outside data sources (e.g., image retrieval and captioning) could possibly improve results on moment localization, both with and without temporal language queries.
Additionally, in Table~\ref{tab:results_prior}, even when the MLLC model can properly localize context, it does not always properly localize temporal sentences indicating that improved temporal reasoning can also improve our results.
We believe our dataset, analysis, and method are an important step towards better moment retrieval models that effectively reason about temporal language.

%% file: acknowledgements.tex
\section*{Acknowledgements}

We thank Anna Rohrbach for helpful feedback.

%% file: supp.tex
In this supplemental, we include an example illustrating how we create TEMPO-TL sentences and include additional qualitative examples.

\section{\datasetname{}-Template Language}

\begin{figure*}[h]
\centering
\includegraphics[width=\linewidth]{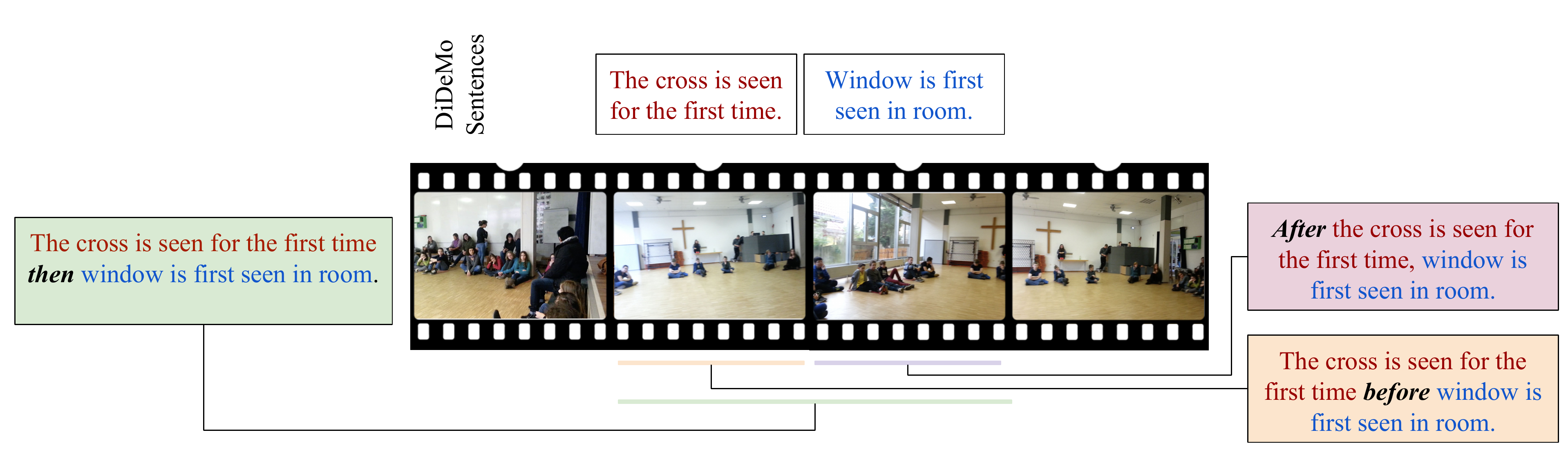}
\caption{\datasetname \ - Template Language (TL).  We use sentence templates for ``before'', ``after'', and ``then'' to transform human provided sentences from the DiDeMo dataset (top) to temporal language queries (left and right).
}
\label{fig:synthetic}
\end{figure*}

Figure~\ref{fig:synthetic} shows examples from our \datasetname{}-TL dataset.
At the top, we show two sentences from the original DiDeMo dataset (``the cross is seen for the first time'' and ``window is first seen in room'').
A ``then'' moment (left, green) is created by concatenating the two adjacent moments from the DiDeMo dataset and combining the sentences using the word ``then'' (``The cross is seen for the first time then window is first seen in room.'').
An ``after'' moment (right, pink) is constructed by referencing the moment which occurs first (``After the cross is seen for the first time'') and then adding the base moment ``window is first seen in room.''
Finally, a ``before'' moment is constructed by concatenating the two adjacent sentences with the word before to form the sentence ``The cross is seen for the first time before window is first seen in room.''

\section{Qualitative Examples on TEMPO-Human Language}

\begin{figure}[t]
\centering
\includegraphics[width=\linewidth]{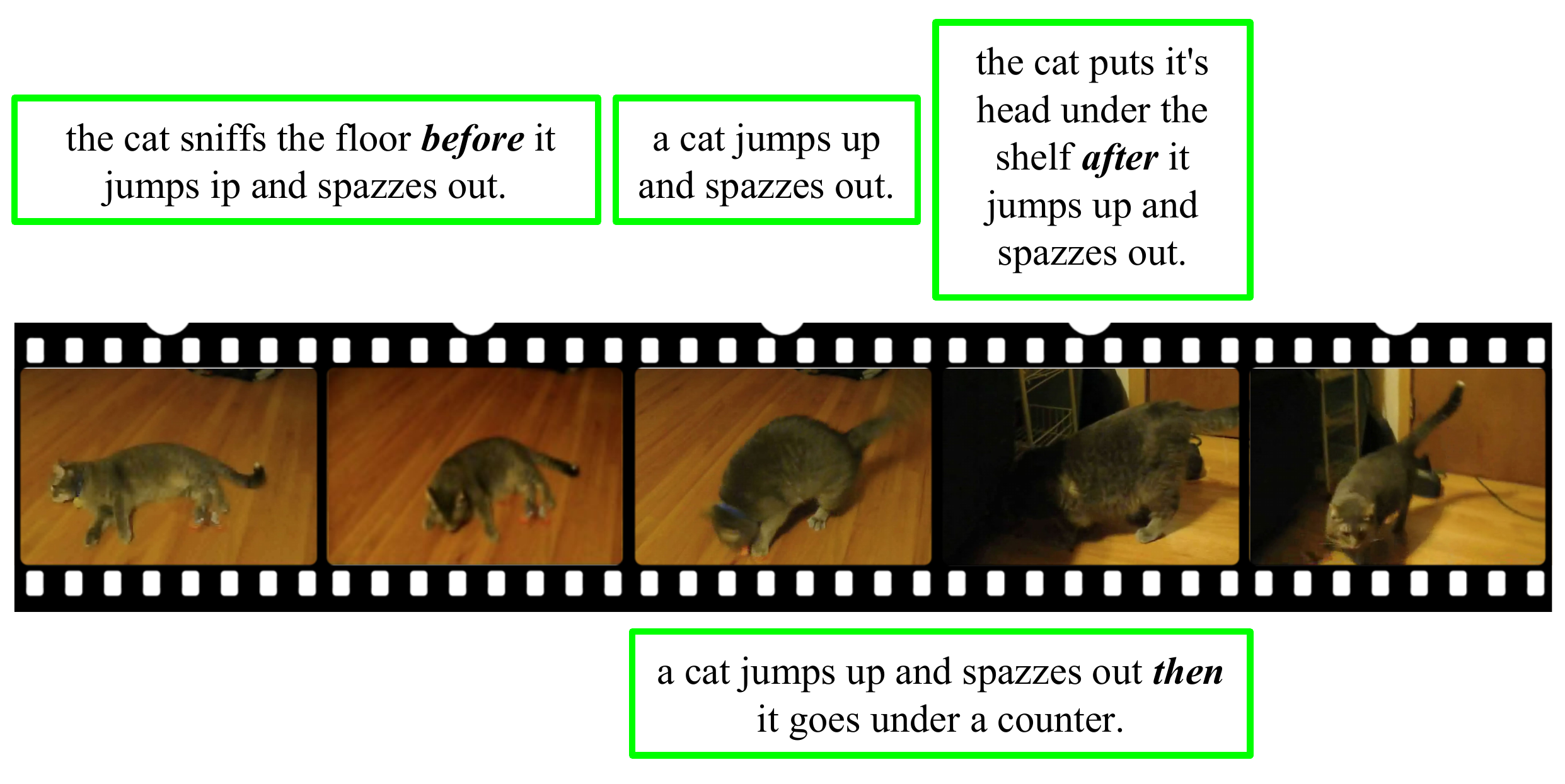}
\caption{Our MLLC model is able to localize original sentences from DiDeMo (a cat jumps up and spazzes out) as well as newly collected sentence from \datasetname{}-HL which include temporal words.
}
\label{fig:many_annos}
\vspace{-5mm}
\end{figure}

\begin{figure}[h]
\centering
\includegraphics[width=0.8\linewidth]{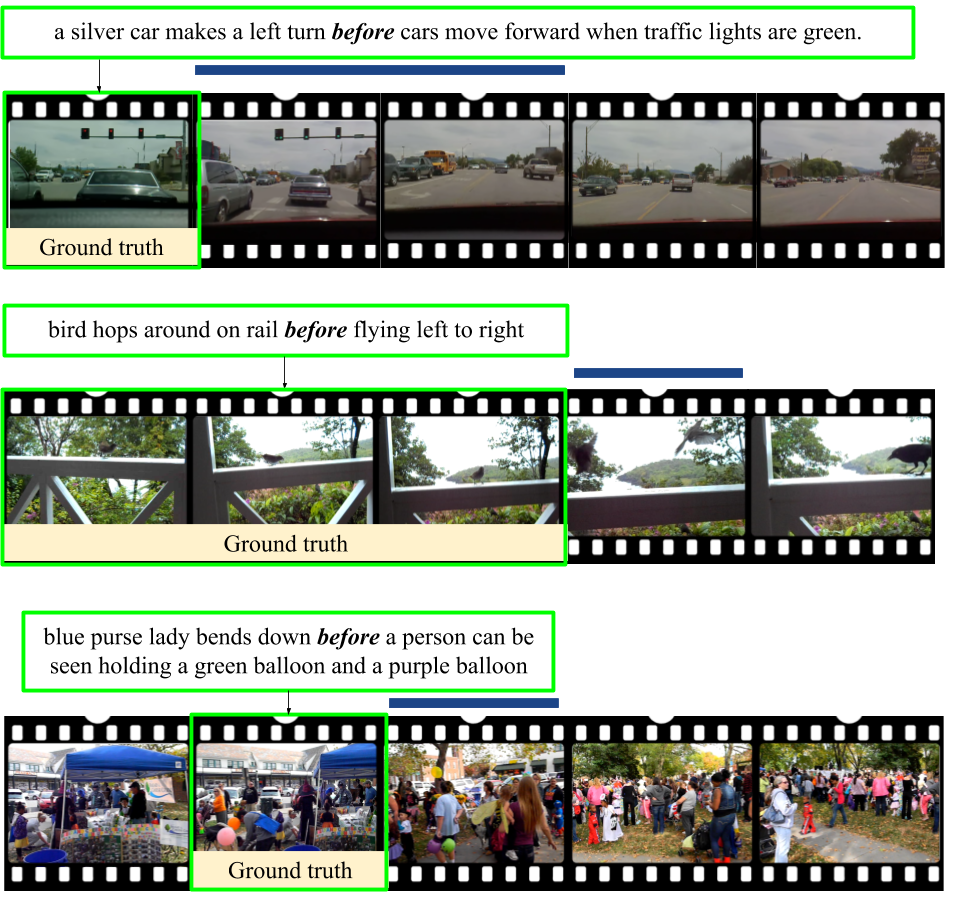}
\caption{\datasetname \ - Human Language (HL).  Localized moments using the temporal word ``before''.  The blue line shows the context considered when localizing the moment, and the correctly predicted moment is highlighted in green.  
}
\label{fig:supp_before}
\vspace{-5mm}
\end{figure}

\begin{figure}[h]
\centering
\includegraphics[width=0.8\linewidth]{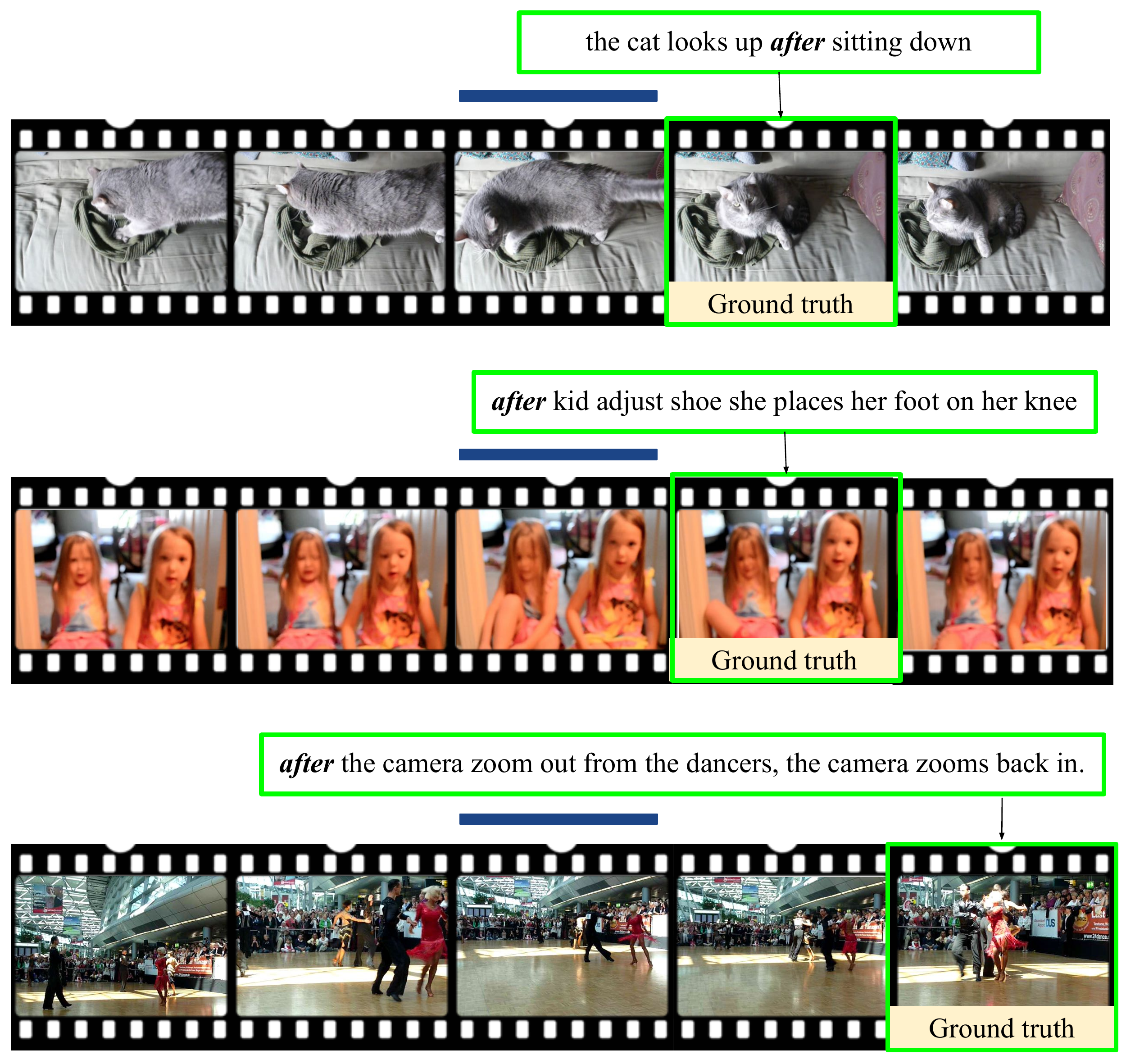}
\caption{\datasetname \ - Human Language (HL).  Localized moments using the temporal word ``after''.  The blue line shows the context considered when localizing the moment, and the correctly predicted moment is highlighted in green.  
}
\label{fig:supp_after}
\vspace{-5mm}
\end{figure}

\begin{figure}[h]
\centering
\includegraphics[width=0.8\linewidth]{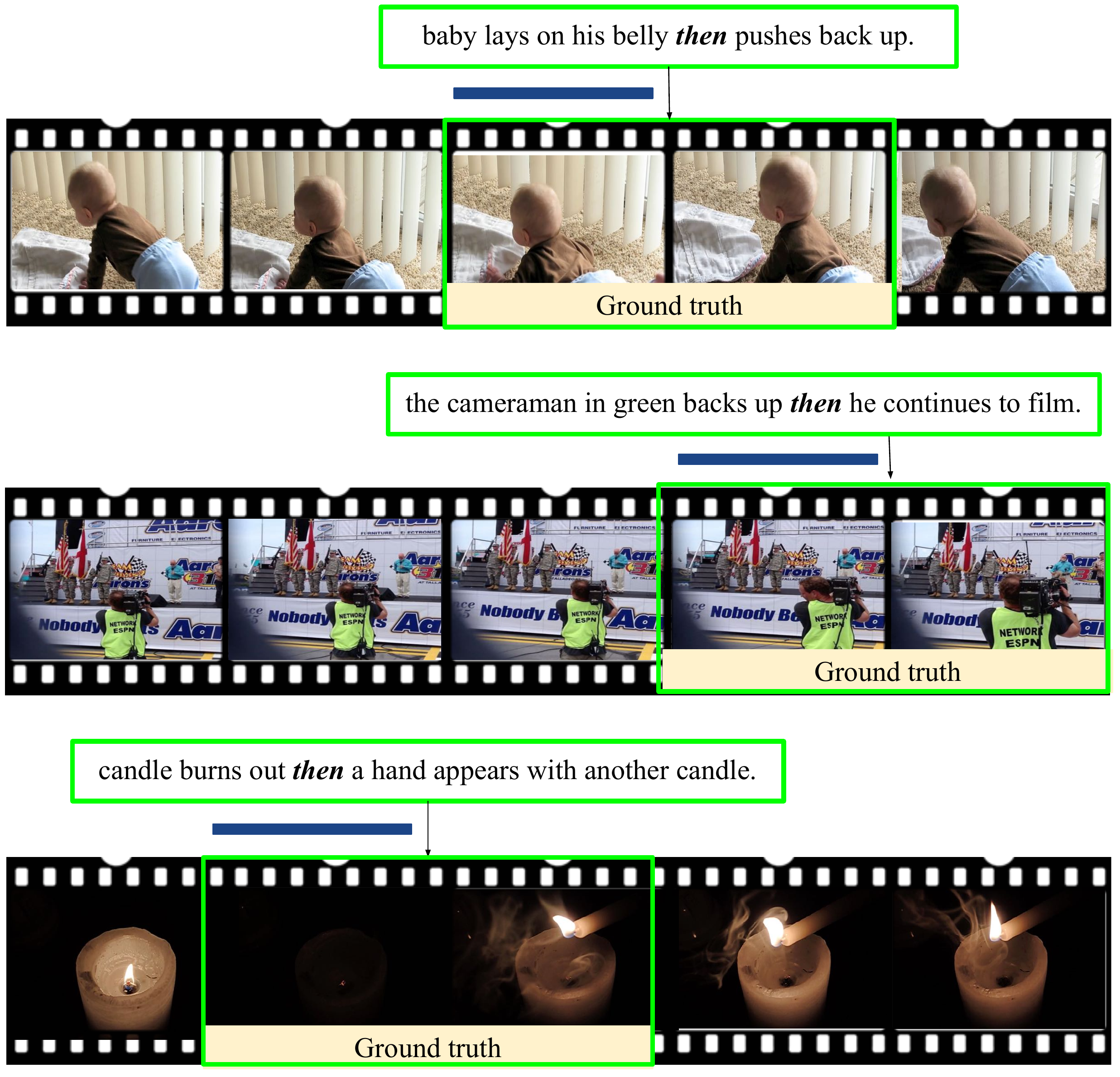}
\caption{\datasetname \ - Human Language (HL).  Localized moments using the temporal word ``then''.  The blue line shows the context considered when localizing the moment, and the correctly predicted moment is highlighted in green. 
}
\label{fig:supp_then}
\vspace{-5mm}
\end{figure}

\begin{figure}[h]
\centering
\includegraphics[width=0.8\linewidth]{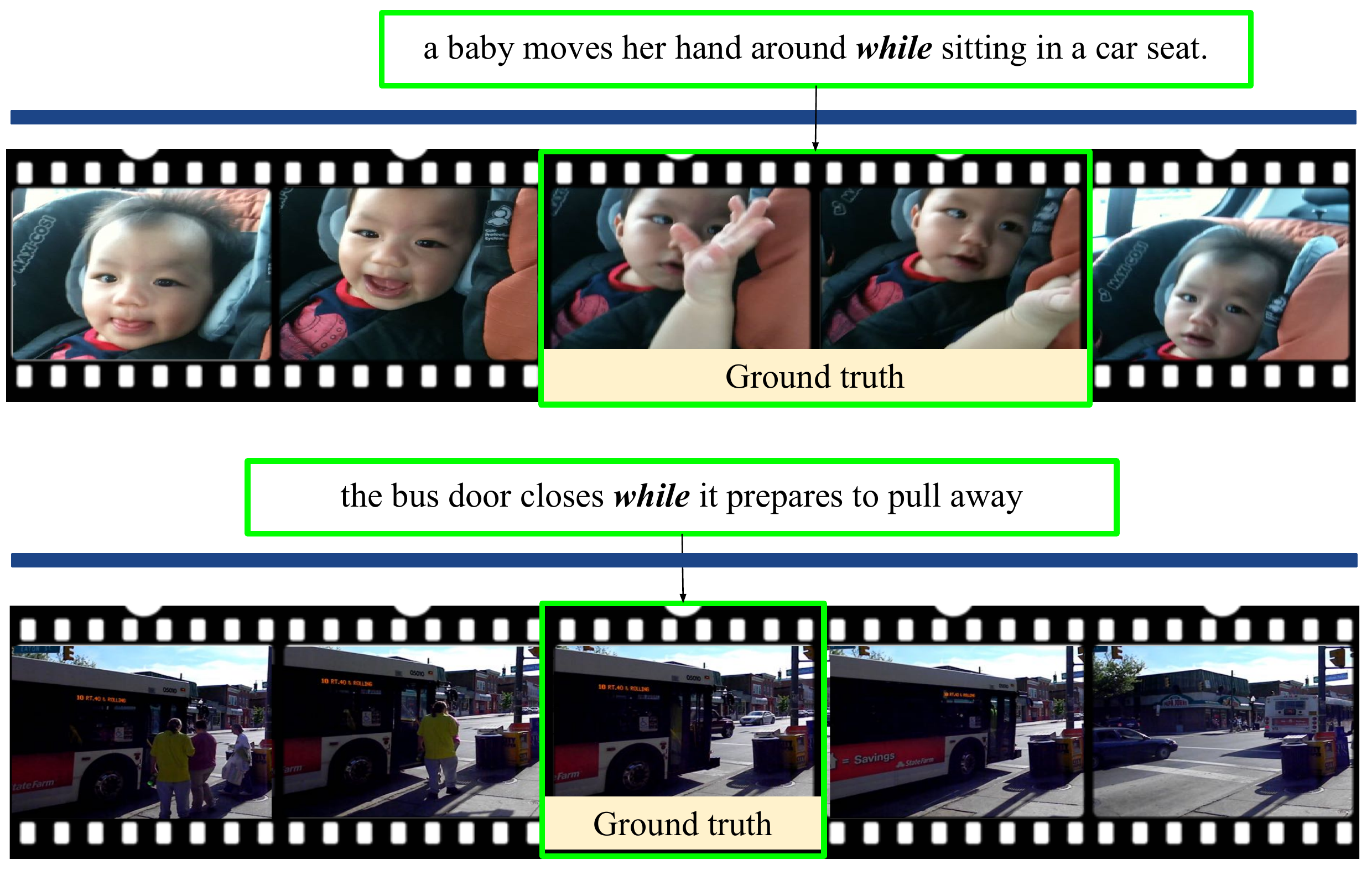}
\caption{\datasetname \ - Human Language (HL).  Localized moments using the temporal word ``while'.  The blue line shows the context considered when localizing the moment, and the correctly predicted moment is highlighted in green. 
}
\label{fig:supp_while}
\vspace{-5mm}
\end{figure}

\begin{figure}[h]
\centering
\includegraphics[width=0.8\linewidth]{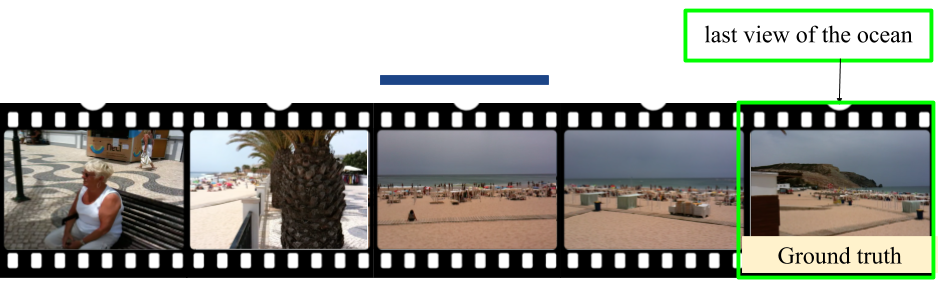}
\caption{\datasetname \ - Human Language (HL).  Localized moment from the DiDeMo dataset.  The blue line shows the context considered when localizing the moment, and the correctly predicted moment is highlighted in green. 
}
\label{fig:supp_didemo}
\vspace{-5mm}
\end{figure}

\begin{figure}[h]
\centering
\includegraphics[width=0.8\linewidth]{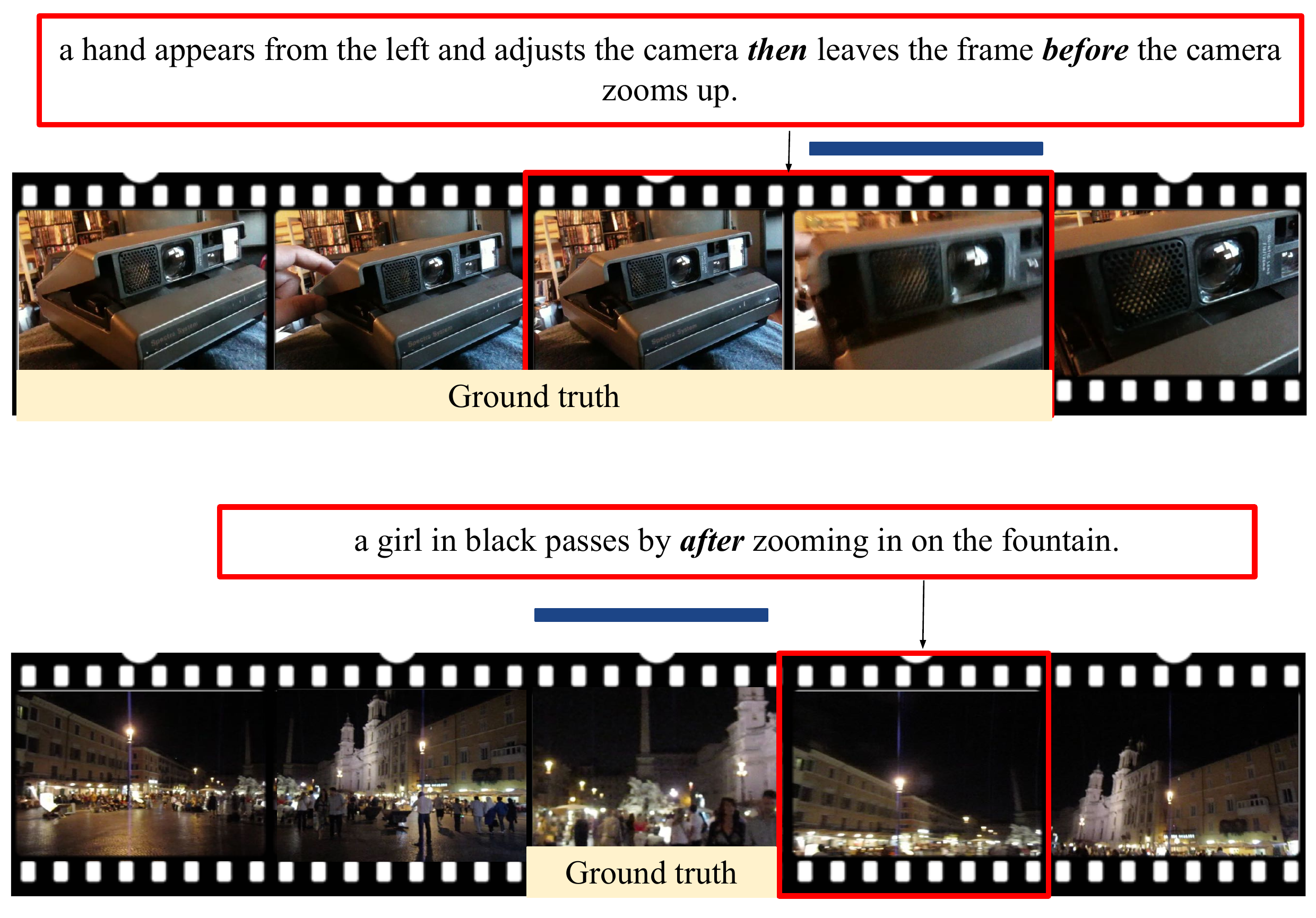}
\caption{\datasetname \ - Human Language (HL).  Example failure cases.  The blue line shows the context considered when localizing the moment, and the (incorrectly) predicted moment is highlighted in red.  In the top example, the context is localized correctly, but the moment is not.  In the bottom example, the temporal relationship between the context and localized moment is sensible, but the localized moment is incorrect.
}
\label{fig:supp_failure}
\vspace{-5mm}
\end{figure}

In this section we show qualitative results for when multiple sentences for the same video are localized properly (Figure~\ref{fig:many_annos}), examples in which our model properly localizes sentence queries (Figures~\ref{fig:supp_before}-\ref{fig:supp_didemo}), and failure cases (Figure~\ref{fig:supp_failure}).

\textit{Comparing Different Temporal Words for the Same Video.}  Figure~\ref{fig:many_annos} shows an example where the original DiDeMo moment is localized correctly (``a cat jumps up and spazzes out'') as well as temporal sentences ``the cat sniffs the floor \textit{before} it jumps ip [sic] and spazzes out'', ``a cat jumps up and spazzes out \textit{then} it goes under the counter'', and ``the cat put it's head under the shelf \textit{after} it jumps up and spazzes out.''  Our MLLC model is able to properly localize each sentence type.  Notice that when the annotators construct temporal sentences, they include pronouns like ``it'' as opposed to repeating the word ``cat'' multiple times.

\textit{Additional qualitative examples.}  In Figures~\ref{fig:supp_before},~\ref{fig:supp_after},~\ref{fig:supp_then},~\ref{fig:supp_while},~\ref{fig:supp_didemo}, and~\ref{fig:supp_failure}, we show additional qualitative examples of the moments localized by MLLC as well as the corresponding context moments.
In general, we observe the context moment is accurately localized and is sensible for each temporal sentence.

Figure~\ref{fig:supp_before} shows correctly localized moments which include the temporal word ``before''.  The top example shows an example in which the context considered by the MLLC model (``cars move forward when traffic lights are green'') consists of multiple GIFs.
Figure~\ref{fig:supp_after} shows correctly localized moments which include the temporal word ``after''.
The bottom example shows an example (``after the camera zoom out from the dancers, the camera zooms back in'') where the localized moment and localized context are not contiguous.
Figure~\ref{fig:supp_then} shows correctly localized moments which include the temporal word ``then''.
In contrast to the temporal words ``before'' and ``after'', to correctly localize ``then'' the chosen context occurs \textit{within} individual moments.
Finally, Figure~\ref{fig:supp_while} shows examples in which ``while'' is used and the context moment corresponds to the global context moment.
Figure~\ref{fig:supp_didemo} shows an example from the original DiDeMo dataset in which global context is not chosen by the MLLC model.
Rather, for the sentence ``last view of the ocean'', the context corresponds to a moment earlier in the video in which the ocean appears.
Because the MLLC model learns to choose context most appropriate for each query, it is not restricted to always using global context.
For sentences like ``last view of the ocean'', choosing context which corresponds to when the ocean is seen earlier in the video is sensible.
This observation may partially explain why the MLLC model does better on the original DiDeMo dataset.

Finally, Figure~\ref{fig:supp_failure} shows interesting failure cases.
Figure~\ref{fig:supp_failure} (top) shows an example where the context was localized properly, but the moment was not correctly localized. 
This could be in part because the query is particularly complex and uses two temporal words (``then'' and ``before'').
Finally, Figure~\ref{fig:supp_failure} (bottom) shows a failure case for the word ``after'' where the context moment is temporally related to the localized moment in a sensible way; for an ``after'' sentence we expect the context to occur \textit{before} the localized moment.

